\documentclass{article}

\usepackage[final]{neurips_2021}
\usepackage[utf8]{inputenc} 
\usepackage[T1]{fontenc}    
\usepackage{hyperref}       
\usepackage{url}            
\usepackage{booktabs}       
\usepackage{amsfonts}       
\usepackage{nicefrac}       
\usepackage{microtype}      
\usepackage{xcolor}         
\usepackage[english]{babel}
\usepackage{amsmath}
\usepackage{ bbold }
\usepackage{enumitem}
\usepackage{wrapfig}
\usepackage{cleveref}
\usepackage{comment}
\usepackage{graphicx}
\usepackage{todonotes}
\usepackage[ruled,vlined,linesnumbered]{algorithm2e} 
\DontPrintSemicolon
\SetAlgoCaptionSeparator{.}
\crefname{algocf}{algorithm}{algorithms}
\Crefname{algocf}{Algorithm}{Algorithms}

\title{\texttt{WAFFLE}: Weighted Averaging for\\ Personalized Federated Learning}

\author{%
  Martin Beaussart \\
  EPFL\\
  \texttt{martin.beaussart@epfl.ch}
  \And
  Felix Grimberg\\
  EPFL\\
  \texttt{felix.grimberg@epfl.ch}
  \And
  Mary-Anne Hartley \\
  EPFL\\
  \texttt{mary-anne.hartley@epfl.ch}
  \And
  Martin Jaggi\\
  EPFL\\
  \texttt{martin.jaggi@epfl.ch}
}

\begin{document}

\maketitle

\begin{abstract}
In federated learning, model personalization can be a very effective strategy to deal with  heterogeneous training data across clients. We introduce \texttt{WAFFLE} (Weighted Averaging For Federated LEarning), a personalized collaborative machine learning algorithm that leverages stochastic control variates for faster convergence. \texttt{WAFFLE} uses the Euclidean distance between clients’ updates to weigh their individual contributions and thus minimize the personalized model loss on the specific agent of interest.
Through a series of experiments, we compare our new approach to two recent personalized federated learning methods---\texttt{Weight Erosion} and \texttt{APFL}---as well as two general FL methods---\texttt{Federated Averaging} and \texttt{SCAFFOLD}. Performance is evaluated using two categories of non-identical client data distributions---concept shift and label skew---on two image data sets (MNIST and CIFAR10). Our experiments demonstrate the comparative effectiveness of \texttt{WAFFLE}, as it achieves or improves accuracy with faster convergence.
\end{abstract}

\newcommand{\NeurIPSinstructions}[1]{ {\color{blue} #1} }

\newcommand{\mathc}{\boldsymbol{c}}
\renewcommand{\c}{$\mathc $}
\newcommand{\mathci}{\mathc_i}
\newcommand{\ci}{$\mathci $}
\newcommand{\mathx}{\boldsymbol{x}}
\newcommand{\x}{$\mathx $}
\newcommand{\mathy}{\boldsymbol{y}}
\newcommand{\mathyi}{\mathy_i}
\newcommand{\yi}{$ \mathyi $}
\newcommand{\istar}{$i^\star$}
\newcommand{\mathS}{\mathcal{S}}
\renewcommand{\S}{$\mathS $}
\newcommand{\mathSet}{\{1,\dots,N\}}
\newcommand{\dmin}{\mathit{dm}}
\newcommand{\dmax}{\mathit{dM}}
\newcommand{\dbot}{d_{i^\star}}

\definecolor{myRed}{RGB}{223, 43, 43}
\definecolor{myOrange}{RGB}{223, 98, 43}
\definecolor{myYellow}{RGB}{223, 219, 43}

\newcommand{\algoHLcolor}{red}
\newcommand{\algoHL}[1]{ {\color{\algoHLcolor} #1 } }

\section{Introduction}
Federated learning (FL) is a collaborative learning technique to build machine learning (ML) models from data distributed among several participants ("agents"). The objective is to generate a common, robust model not by exchanging the data between agents, but rather through exchanging parameter updates for the common model that all agents share at a certain frequency. This technique addresses some major problems of centralized data-sharing approaches, can enable data privacy and security and is therefore attractive to many applications with sensitive data.

We consider centralized FL, in which a single server orchestrates the execution of the algorithms. Each iteration of learning can thus be divided into four main steps. Firstly, the central server sends the current model to all agents. Secondly, each agent trains the model with their data and, thirdly, sends the updated model parameters back to the central server. Finally, the central server aggregates these results and generates a new global model. \texttt{WAFFLE} intervenes in this last step: instead of a global one-size-fits-all model, the aggregation produces a personalized model for each specific agent.

\subsection{Control Variates for Heterogeneous Data}
\label{subsection:Intro_stochastic_control}
Federated learning develops a common model for all agents, typically under the assumption that data is \emph{independent and identically distributed (IID)} across agents. In real-world applications, this assumption rarely holds, and thus the convergence rate and final performance of FL algorithms like \texttt{Federated Averaging} (\texttt{FedAvg}) can vary significantly \citep{mcmahan2017fedavg,karimireddy2020scaffold}.
\citet{karimireddy2020scaffold} introduce \texttt{SCAFFOLD}, which uses stochastic control variates (SCV) to tackle the \emph{client-drift} that results from non-IID-ness across agents. Each agent maintains an SCV estimating the sum of all other agents' gradients, and uses it to ``correct'' each step of stochastic gradient descent (SGD). Compared to \texttt{FedAvg}, \texttt{SCAFFOLD} converges faster and towards a better model.
\texttt{WAFFLE} extends \texttt{SCAFFOLD} to collaboratively train a personalized ML model.

\subsection{Model Personalization} 
A complementary solution for dealing with non-IID data is to train a unique (personalized) model for each agent, rather than a single global model. 
Recently, there have been several key works on personalization in federated learning, see e.g. \citet{kulkarni2020survey} for an overview. Some methods of adapting global models for individual agents are summarized in \citet{kairouz2019_FL_advances_open_problems}, such as local fine-tuning \citep{deng2020apfl,kairouz2019_FL_advances_open_problems}, multi-task learning \citep{Smith2017MOCHA}, and model-agnostic meta-learning (MAML) \citep{Finn2017,fallah_personalized_MAML_2020}. 

The model personalization method most similar to \texttt{WAFFLE} is \texttt{Weight Erosion} (\texttt{WE}) \citep{grimberg2020weight}.
It is conceptually related to local fine-tuning, which consists of training a common global model and subsequently personalizing it for each agent by performing a small number of SGD steps on the agent's local data (resulting in one model per agent) \citep{kairouz2019_FL_advances_open_problems}.
\texttt{WE} achieves a smoother and more differentiated
transition from global to local training
 by using a \emph{weighted average} of the agents' gradients to update the server model at each communication round.
Thus, the server model after running \texttt{WE} once is personalized for one specific agent---called \emph{user} or \emph{Alice}.\footnote{
   The authors motivate this setting, in which all other agents contribute selflessly to Alice's objective, by constructing a fictive scenario of cross-silo FL across hospitals.
   Alternatively, for cross-silo FL, \texttt{WE} (or \texttt{WAFFLE}) can be run several times in parallel to obtain personalized models for all agents.
}
Like \texttt{WE}, \texttt{WAFFLE} uses aggregation weights derived from the Euclidean distance between gradients to update the global model, but it combines this approach with the use of SCVs (\Cref{subsection:Intro_stochastic_control}).
These different approaches for transitioning from global to local training are illustrated in \Cref{evol_weight} (\Cref{section: Benchmarking personalized collaborative learning methods.}).

Instead of interpolating between gradients at each round, a slightly different approach consists of interpolating between a local and global model \citep{grimberg2021optimal,Donahue2020stable_coalitions,deng2020apfl,zhang2021personalized}. For instance, \citet{deng2020apfl} propose the adaptive FL algorithm \texttt{APFL} that achieves personalization by learning a global model and a set of local corrective models with which to interpolate.
Another related approach is \texttt{personalized MAML}, where a global model is optimized with respect to (w.r.t.) the \emph{loss after local fine-tuning} on each agent's local data \citep{Finn2017,fallah_personalized_MAML_2020}.
Finally, regularization can be used (instead of weighted averaging) to  mix global and local training \citep{Smith2017MOCHA,li_ditto_FL_2021}.
For instance, \citet{li_ditto_FL_2021} recently showed very promising results with \texttt{Ditto}, where the local model's training is split across all communication rounds and regularized by its distance from the current global model.

\subsection{Benchmark}
 While several categories of non-IID data have been identified by \citet{kairouz2019_FL_advances_open_problems}, 
%
%
we focus only on label skew and concept shift:
\begin{itemize}
    \item Label skew: The distribution of labels varies between agents, but the "true" classification function is the same for all agents. This is, so far, the type of non-IID-ness most commonly used in personalized FL benchmarks \citep{li_ditto_FL_2021,fallah_personalized_MAML_2020,deng2020apfl}.
    \item Concept shift: The mapping from the features to the label varies across agents. We include concept shift to model naturally partitioned data sets like those used by \citet{Smith2017MOCHA}.
\end{itemize}


\subsection{Contributions}
\begin{itemize}
    \item We present a new personalized collaborative ML algorithm named \texttt{WAFFLE} (Weighted Averaging For Federated LEarning), which builds on the existing methods \texttt{SCAFFOLD} and \texttt{Weight Erosion} (\texttt{WE}) by using stochastic control (SC) and weighted gradient aggregation. To the best of our knowledge, this is the first method that uses SC in model personalization.
\item We build an image classification benchmark with label skew and concept shift. We then use it to compare \texttt{WAFFLE} against baselines (\texttt{Federated Averaging} and \texttt{Local}), its parent methods (\texttt{SCAFFOLD} and \texttt{WE}), and against other state-of-the-art personalized FL methods (\texttt{APFL}), thus contributing to further empirical evaluation of these methods.
\item Finally, we show that in most cases, \texttt{WAFFLE} converges faster than its competitors \texttt{WE} and \texttt{APFL} and matches or improves their accuracy, while requiring less hyperparameter tuning.
\end{itemize}

\section{The \texttt{WAFFLE} Algorithm}

To introduce \texttt{WAFFLE}, we will first summarize the known \texttt{SCAFFOLD} FL algorithm on which it is based, and then detail how personalization can be achieved by weighted averaging between agents, for suitable weight choices. \texttt{SCAFFOLD} maintains an estimate of what all agents are learning, called a \textit{control variate}, so that agents can simulate having all the data and estimate the direction in which they should update their gradient. Thus, it is more efficient and addresses the problem of \emph{client drift}. Indeed \citet{karimireddy2020scaffold} prove that \texttt{SCAFFOLD} converges to the globally optimal model instead of converging to a weighted average of local models and does so much faster and more accurately than \texttt{FedAvg} in the presence of inter-agent heterogeneity. 

\subsection{Personalization Using Weighted Update Aggregation}
\texttt{WAFFLE} is essentially a personalized version of \texttt{SCAFFOLD}, where the goal is no longer to get a global model for all agents but a personalized model for one particular agent, (\textit{Alice}). The idea is to start from global training (\texttt{SCAFFOLD}) and to gradually move to local training. \texttt{SCAFFOLD} computes the gradient for the server model ($\Delta \mathx$) and the \textit{control variate} $\Delta \mathc$ (the assumption of what other agents are learning) by averaging each agent’s update $\Delta \mathyi$ and local \textit{control variates} \ci\, at each round \citep[Algorithm 1, Line 16]{karimireddy2020scaffold}.
It is at this step that \texttt{WAFFLE} intervenes: Instead of averaging all agents at each round, \texttt{WAFFLE} assigns a weight $\alpha_i$ to each agent and computes $\Delta \mathx$ and $\Delta \mathc$ by a weighted combination as per \Cref{eqn:weighted update aggregation}. Changes w.r.t. \texttt{SCAFFOLD} are highlighted in red:
\begin{equation}\label{eqn:weighted update aggregation}
\left(\Delta \mathx,
\Delta \mathc \right) \leftarrow \sum_{i \in \textcolor{\algoHLcolor}{\mathSet}} \textcolor{\algoHLcolor}{\alpha_{i}} \left(\Delta \mathyi,\Delta \mathci \right)
\end{equation}

For $N$ agents, we recover \texttt{SCAFFOLD} (global training) by setting all weights equal to $\frac{1}{N}$. In contrast, we recover local training by setting all weights to $0$, except the weight of Alice.

\begin{figure}[htp]
    \centering
    \includegraphics[scale=0.12]{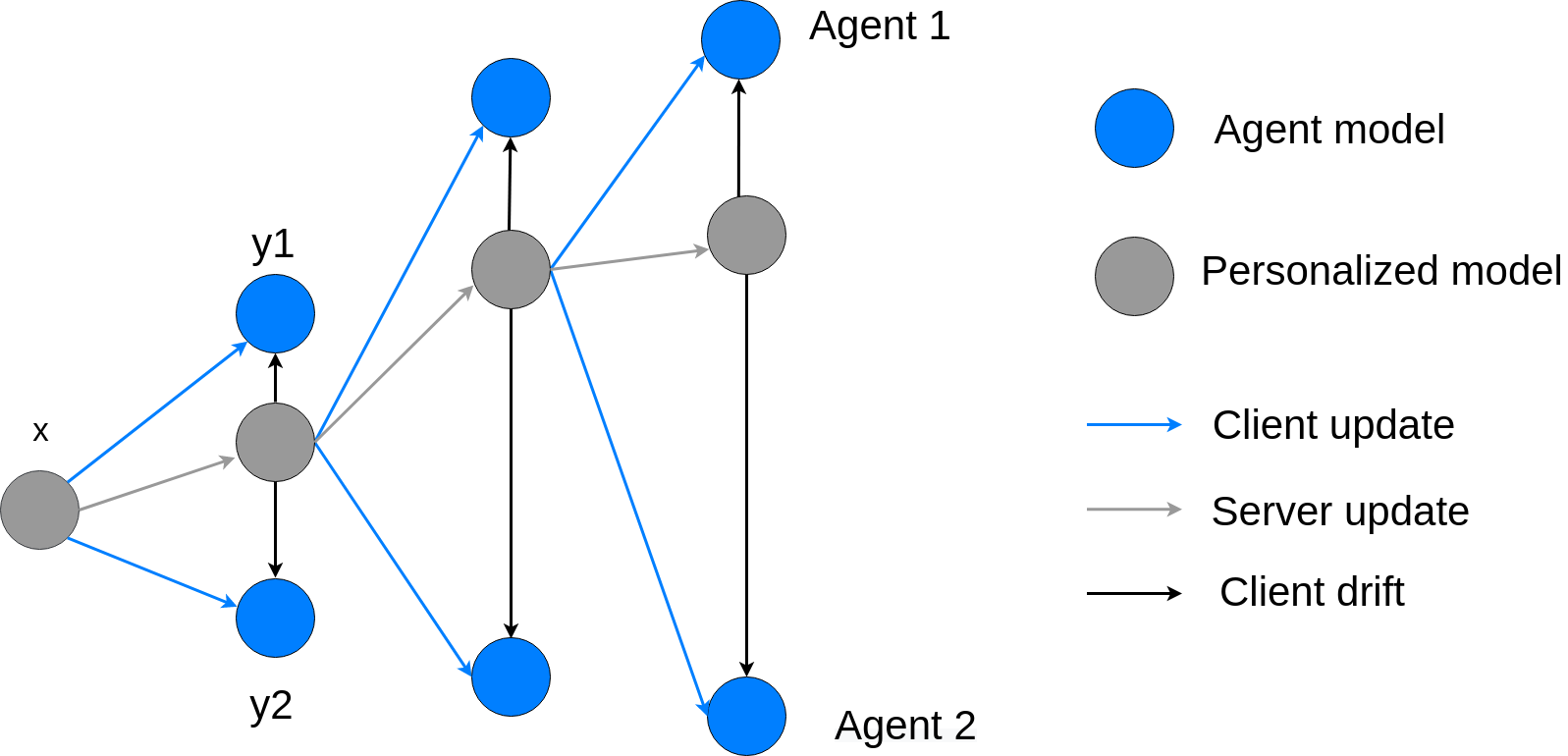}
    \caption{Evolution of the personalized model and agent models on \texttt{WAFFLE} in a three-rounds scenario where agent 1 has a higher weight than agent 2. }
    \label{client_drift} 
\end{figure}

\texttt{WAFFLE} is based on a smooth transition from global to local training. To do this, the weight of each agent is updated according to the degree of personalization desired for the round. Just as \texttt{SCAFFOLD} uses the control variates to converge to the global model for the union of all agents' datasets, \texttt{WAFFLE} uses them to converge to a personalized model based on  a weighted subset of agents (\Cref{client_drift}). Each agent is included and weighted in this subset based on the current degree of personalization and whether their dataset is sufficiently similar to Alice's, as measured by the distance between their gradients at each round.

The weight of each agent at a given round depends on the distance between its and Alice's gradient. As the method uses SGD (i.e., stochastic gradients), the agent weights are also subject to stochastic noise. Indeed, at some rounds, some agents may have an atypical gradient resulting in a bad weight, even if their contribution is still needed for the global model. Therefore, we average the current weight with the last two weights to smooth out strong random effects (line 10 of \Cref{algo:omega}).


To compare each agent with Alice, we use the Euclidean distance between gradients. Simply, we want an agent with a small distance to have a higher weight than an agent with a larger distance.
Like \texttt{SCAFFOLD}, we have a central server that coordinates all the agents. At each starting round it gives the current personalized model and control variate to all agents for training. When the agents finish training the model with their data, they send the updated model and their local control variate back to the server. The server then calculates the Euclidean distance of each update from the selected agent and modifies the weight assigned to each agent accordingly. Finally, it calculates the new personalized model and the new control variate (\Cref{eqn:weighted update aggregation}).
The implementation of \texttt{WAFFLE} can be seen in the modified version of \texttt{SCAFFOLD} (\Cref{algo:waffle}, \Cref{appendix:figures tables algorithms}) where changes w.r.t. to the original code are written in red ink. Agent weights are obtained by CalcWeight (\Cref{algo:omega}), which is explained in the next subsection.

\subsection{Computing Agent Weights for \texttt{WAFFLE}}\label{sec:selectweights}
\texttt{WAFFLE} uses a new data-dependent approach to select the contribution weight $\alpha_{i}^r$ of agent $i$ at each round $r$. It relies on hyperparameters $\Omega$ and $\Psi$: two functions defining the degree of personalization for each round $r$ (where a value of $1$ corresponds to global training and $0$ is personalized (local) training). As explained later, we reduce the functions $\Omega$ and $\Psi$ to a single, positive real-valued hyperparameter for ease of tuning.

The weight is calculated according to \Cref{Eq1,Eq2}, where\vspace{-2mm}
\begin{itemize}[itemsep=1pt]
    \item agent $i^\star$ is Alice
    \item $d_{i}$ is the distance between the gradients of agents $i$ and $i^\star$
    \item $\dmax$ is the largest such distance: $\dmax \leftarrow \underset{i}{\max} \ d_i$ \vspace{-1mm}
    \item Analogously, $\dmin$ is the smallest such distance (excluding $d_{i^\star}$)
    \item $\Omega(r)$ and $\Psi(r)$ denote the degree of personalization for round $r$
\end{itemize}
The weight of an agent is based on the distance between its gradient and the gradient of agent $i^\star$, but also on the distribution of the distances of the agents. The distance of $i^\star$ would always be 0, we want to change it to a similar value compared to the distances of other agents. The further away Alice's distance is from the others, the less weight we assume the agents have compared to Alice.
We need a ``distance'' $\dbot$ for Alice in  \Cref{Eq2}. Thus, 
\Cref{Eq1} serves to assign her such a distance based on the hyperparameter $\Omega$. When $\Omega \approx 1$ (i.e., more global training), Alice's assigned ``distance'' ($\dbot$) is close to the next smallest distance $\dmin$, resulting in similar weights for these two agents. The more $\Omega$ decreases towards 0, the smaller ($\dbot$) becomes compared to all other agents' distances, resulting in a large weight for Alice (i.e., more local training). 
\begin{equation}\label{Eq1}
\dbot \leftarrow \dmin \cdot \Big( 1 - \textcolor{blue}{\frac{\dmax - \dmin}{\dmax}} (1 - \Omega(r) ) \Big)
\end{equation}
\begin{equation}\label{Eq2}
 \alpha_i^r \leftarrow   \max\Big\{ \Psi(r)    - \frac{ d_{i} - \dbot}{\dmax - \dbot}
 ,\, 0\Big\}
\end{equation}

\Cref{Eq2} serves as a threshold of inter-agent utility. To pass the threshold, the gradient of agent $i$ must be closer than the other agents' gradients (based on the range of distances). With $\Psi$ close to 0 (i.e., more local training), only a small fraction of the range between the minimum and maximum distance will map to a non-zero weight. Thus, learning is concentrated in a subset of agents with the most similar gradients.

Using the two functions $\Omega$ and $\Psi$, we can define different strategies for \texttt{WAFFLE}. We could, as an example, shift rapidly toward zero and thus drastically limit larger steps in global training in the initial phase of personalization.

As \Cref{algo:omega} would otherwise have $2R$ positive real-valued parameters to tune (values of $\Omega$ and $\Psi$ for each round), we restrict ourselves to $\Omega (r) = \Psi (r) = $ a single-parameter function of $r$. However, more methods should be investigated to see the full potential of \texttt{WAFFLE} in different scenarios.

In this paper, we use a sigmoid function to give the general training a gradual slope towards smaller values \Cref{waffle_s_omega} and \Cref{omega_curve}.
The function has a parameter $\Delta\Omega$ that controls the slope, a higher value will steepen the slope and thus move faster towards the local training. According to our experiments, a good value for $\Delta\Omega$ is around 3.2, irrespective of the model used.
With this value, \texttt{WAFFLE} transitions to mostly local SGD after $70-90\%$ of the total number of rounds, depending on the similarity of the agents' gradient to that of Alice. A simple modification to delay or accelerate the local learning time is to move the $\Omega$ function horizontally by adding a value to~$r$.\vspace{-2mm}
\begin{equation}\label{waffle_s_omega}
\begin{aligned}
\Omega(r) = \Psi(r) = \frac{1}{1+e^{\Delta\Omega\cdot(r/(R/2)-1)})}
\end{aligned}
\end{equation}

\begin{algorithm}[!htbp]
    \SetKwInOut{Input}{input}\SetKwInOut{Output}{output}\SetKwInOut{HyperPar}{hyperparameter}
    \caption{CalcWeights-$\Omega$}
    \label{algo:omega}
    \Input{ $N$, $R$, \istar, $r$, $\{ \Delta \mathyi \}$, $\mathbf{\alpha}^{r-1}$, $\mathbf{\alpha}^{r-2}$}
    \HyperPar{ Functions $\Omega$ and $\Psi$ from $ \{1, \dots, R \}$ to $\mathbb{R}_{>0}$}
    \Output{ Weight vectors $\bar{ \mathbf{\alpha}}^{r}, \mathbf{\alpha}^r$}

    \ForEach{  \emph{agent} $ i \in \mathSet $ }{
        \tcp{Compute the Euclidean distance between updates of agents $i$ and \istar.}
        $ d_{i} \leftarrow  \left \| \Delta \mathyi - \Delta \mathy_{i^\star} \right \|_2 $ \tcp*{$\dbot = 0$}
    }
    $ \dmax \leftarrow \underset{i \neq i^\star }{\max} \ \  d_{i} $, \qquad
    $ \dmin \leftarrow \underset{i \neq i^\star }{\min} \ \   d_{i} $ \;
    $ \dbot \leftarrow \dmin \cdot \left( 1 - \textcolor{blue}{\frac{\dmax - \dmin}{\dmax}} (1 - \Omega(r) ) \right) $
    \tcp*{$0 \leq \dbot \leq \dmin $ if $\Omega(r) \in [0,1]$}
    \ForEach{  \emph{agent} $ i \in \mathSet $ }{
        $ \alpha_i^0 \leftarrow   \max\big\{ \Psi(r)    - \frac{ d_{i} - \dbot}{\dmax - \underset{i}{\min} \ d_i} , 0\big\}$ \;
        \If{$r \geq 0.95 R$}{
            $\alpha_i^0 \leftarrow 0$ if $i\neq i^\star$ else $1$ \;
        }
    }
    $\mathbf{\alpha}^0 \leftarrow \frac{1}{\sum \alpha^0_i} (\alpha^0_1, \dots, \alpha^0_N )^\top$ \;
    $\bar{ \mathbf{\alpha}}^{r} \leftarrow \frac{1}{3} (\mathbf{\alpha}^{-2} + \mathbf{\alpha}^{-1} + \mathbf{\alpha}^0)$ \;

\end{algorithm}

\section{Benchmarking Personalized Collaborative Learning Methods}
\label{section: Benchmarking personalized collaborative learning methods.}

To compare the performance of \texttt{WAFFLE} with the state of the art, we evaluate it against two recent personalized FL methods---\texttt{APFL} \citep{deng2020apfl} and \texttt{Weight Erosion (WE)} \citep{grimberg2020weight}---as well as two global methods---\texttt{Federated Averaging} \citep{mcmahan2017fedavg} and \texttt{SCAFFOLD} \citep{karimireddy2020scaffold}---and finally, local training using only Alice's dataset (\texttt{Local}). \texttt{SCAFFOLD} and \texttt{WE} were selected to validate whether \texttt{WAFFLE} outperforms the methods it builds upon. We included \texttt{APFL} in the benchmark as an unrelated personalized FL method which has an open-source implementation. The benchmark consists of training standard image classification networks on standard IID and non-IID image datasets described below. 

Two 10-class image classification tasks are used based on either the MNIST or CIFAR10 datasets. To generate confidence estimates, we run the MNIST benchmark with five different seeds. However, using five seeds for CIFAR10 would be prohibitive due to computational limitations.

\begin{wrapfigure}{r}{0.45\textwidth}
    \centering
    \includegraphics[scale=0.12]{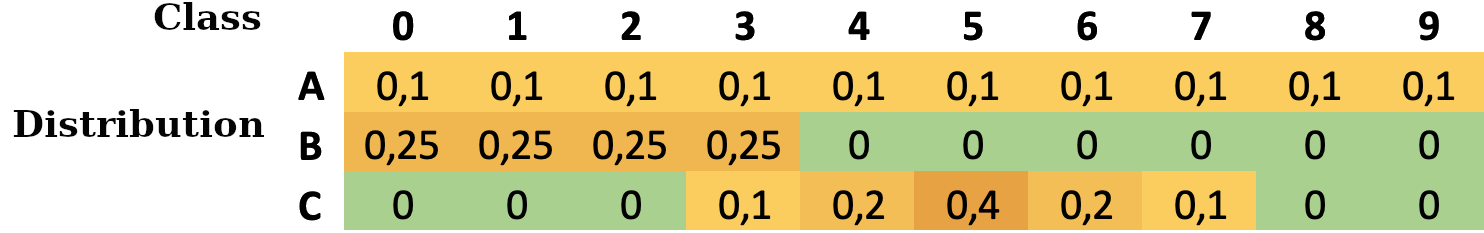}
    \vspace{-3mm}
    \caption{Repartition of labels between all agents for each distribution benchmarked}
    \label{table_distribution}
    \vspace{-2mm}
\end{wrapfigure}

\textbf{Distributions and Models.}
We test five distributions (A,B,C,A*,B*), the first one (A) is IID and serves as a baseline. A classic label skew distribution (B) was selected for consistency with several other benchmarks \citep{Collins2021,deng2020apfl,McMahan2016}, where only a small number of agents have samples from any given class (uniformly distributed amongst these agents). Distribution C exhibits more nuanced label skew, where the labels are not uniformly distributed among the agents. Instead, each agent will have 40\% of one label, 20\% of two labels, and 10\% of two other labels. 
We also present two other distributions---concept shift on the IID distribution (A*), and concept shift on top of classic label shift (B*). Concept shift is achieved by randomly swapping the class labels of all agents, except for Alice. We specify each distribution as a list of ten numbers that sum to one (see \Cref{table_distribution})). For each agent $i$, we move the list to the right $i$ times to get non-IID data, e.g.  with distribution B, agent 0 will only have labels 0 to 3, agent 1 labels 1 to 4, etc.

For MNIST we use a LeNet-5 model \citep{LeCun1989} and for CIFAR10 we use DLA \citep{Yu2017}.

\begin{figure}[htp]
    \vspace{-3mm}
    \centering
    \includegraphics[scale=0.27]{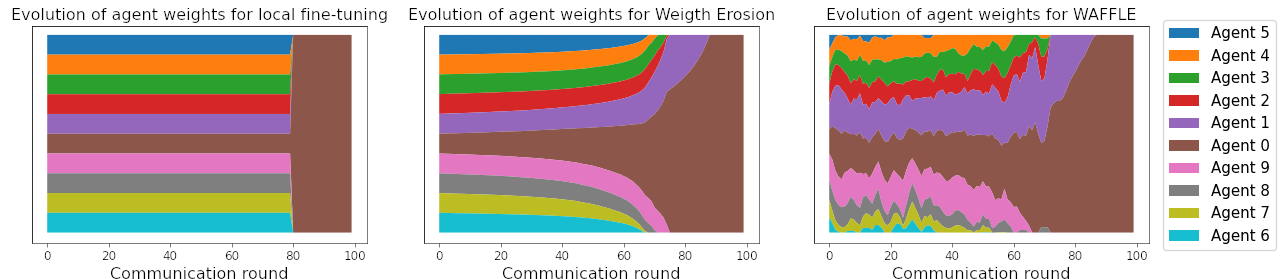}
    \caption{Evolution of weight for personalized FL with the distribution C on MNIST (seed = 1)}
    \label{evol_weight}
\end{figure}

\textbf{Hyperparameters.}
We use an SGD optimizer with a learning rate of 0.1 for MNIST and 0.01 for CIFAR10. Following \citep{Yu2017} we add a momentum 0.9 and a weight decay 5e-4 for CIFAR10 . Concerning the batch size, it is set to 32 for MNIST and 128 for CIFAR10. Except for the \texttt{APFL} method, we define the hyperparameters in a principled manner (based on the expected behaviour of the algorithm). As \texttt{WE} and \texttt{WAFFLE} use a weight system, we attempt to set the parameters in such a way that both algorithms reach fully local learning at about the same time. In practice, this was difficult to achieve as some hyperparameters are very sensitive. Fully local learning is obtained when all weights reach zero except for Alice (Agent 0), who has a weight of one. \Cref{evol_weight} shows an example of the evolution of the weights. Alice (agent 0) is placed in the centre. For the hyperparameters of \texttt{APFL}, we use the same method mention in the paper by \citep{deng2020apfl}, and adaptively update the hyperparameter to guarantee the best generalization performance.\\

The tuned hyperparameters are listed in \Cref{table:parameters} (\Cref{appendix:figures tables algorithms}). Out of the two methods, \texttt{WAFFLE} is the easiest to parametrize, as the hyperparameter does not change in contrast to the weighting system used by \texttt{WE}. This is because \texttt{WAFFLE} can work very well without tuning, as opposed to \texttt{WE} which requires hyperparameter tuning to perform well. As explained in \Cref{sec:selectweights}, we purposefully restrict \texttt{WAFFLE} to a single hyperparameter, $\Delta\Omega$ (\Cref{waffle_s_omega}) and we use the same value for all experiments. Nevertheless, we show in the result section that \texttt{WAFFLE} is able to obtain a very good accuracy even without tuning.

\section{Results}
\label{section: Results}
We present the results of our benchmarking in \Cref{summary benchmark table} below. For each method and each distribution, it shows the best accuracy reached at any of the 100 epochs. In the MNIST section, results are averaged across five random seeds and listed along with the standard deviation as $avg\pm std$.

\begin{wrapfigure}{r}{0.45\textwidth}
    \centering
    \vspace{-3mm}
    \includegraphics[scale=0.3]{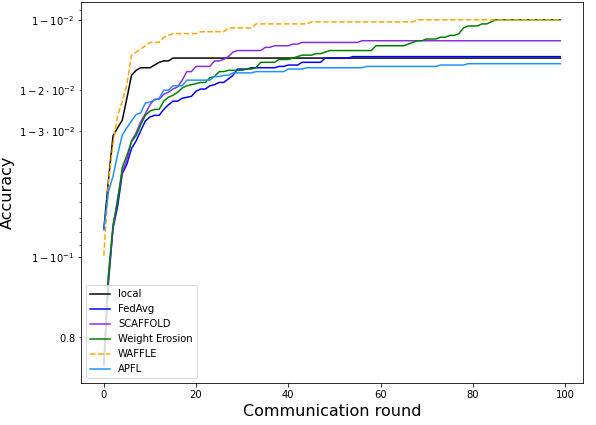}
    \vspace{-2em}
    \caption{Evolution of accuracy with distribution C on MNIST, average over five seeds of the best accuracy obtained up to each turn.}
    \label{fig:evol_acc}
    \vspace{-1mm}
\end{wrapfigure}

To see the evolution of the accuracy, we provide an example in \Cref{fig:evol_acc}. The evolution for each experiment is shown in the appendix (see \Cref{acc_mnist,acc_cifar}).

\textbf{MNIST.}
On MNIST, \texttt{WAFFLE} and \texttt{WE} outperform the other methods for the label skews (B, C), but not for IID distribution (A). The difference between \texttt{WAFFLE} and \texttt{WE} lies in the speed of convergence: As expected \texttt{WAFFLE} converges faster thanks to the use of SCVs.
\\
For the concept shift distributions (A*, B*), all personalized FL methods improve accuracy compared to global FL methods. Among all personalized FL methods, \texttt{WE} performs slightly better in this setting. Here, \texttt{WAFFLE} is not more efficient than \texttt{WE} (time taken to reach convergence). This may be related to the fact that only small fractions of agents are useful in this distribution, so until we reach this critical point, both algorithms are bound to the same slow increase until a threshold. Regarding \texttt{APFL}, the method also performs better than global FL methods except for the IID distribution (A), but it struggles to achieve a better result than the other two personalized FL methods and \texttt{Local}.

\textbf{CIFAR10.}
Concerning CIFAR10, the lack of other seeds makes us less confident about the interpretation of our results, however, the results are coherent with our expectations. For example, like with MNIST, global FL methods outperform personalized methods on the IID distribution and present a significant improvement compared to \texttt{Local}. Distribution B and C (label skew) produce results close to those of MNIST, where \texttt{WAFFLE} and \texttt{WE} outperform others. Concerning concept shift, personalized FL methods are still a great improvement compared to global FL or \texttt{Local}. The accuracy of \texttt{WAFFLE} and \texttt{WE} is approximately the same but as for MNIST we don’t have a noticeable difference between the speed of convergence to best accuracy. For \texttt{APFL}, we get similar results as MNIST, \texttt{APFL} performs better than the global FL methods, but this time it is better than \texttt{Local}. Unfortunately, compared to the other two customized methods, it performs much worse.

\begin{table}[!htbp]
  \caption{Comparison of average local test accuracy and standard deviation of different algorithms given different data distributions.}
  \label{summary benchmark table}
  \centering
  \begin{tabular}{p{1.5cm} p{0.9cm} p{0.9cm} p{1.6cm} p{0.9cm} p{1.3cm} p{1.9cm}}
    \toprule
    \multicolumn{7}{c}{Algorithms}                   \\
    \cmidrule(r){2-7}
     & \texttt{Local} & \texttt{FedAvg} & \texttt{SCAFFOLD} & \texttt{WE} & \texttt{WAFFLE} & \texttt{APFL} \\
    MNIST \\
    \midrule
    IID (A) & $96.88\%$ $\pm 0.47 $ &$98.99\%$ $\pm 0.08 $ &$\mathbf{99.03}\%$ $\pm 0.08 $ &$99.01\%$ $\pm 0.16 $ &$98.93\%$ $\pm 0.1 $ &$97.55\%$ $\pm 0.13 $ \\

    Distr. B & $99.38\%$ $\pm 0.09 $ &$98.97\%$ $\pm 0.18 $ &$99.24\%$ $\pm 0.17 $ &$\mathbf{99.69}\%$ $\pm 0.07 $ &$\mathbf{99.69}\%$ $\pm 0.03 $ &$99.36\%$ $\pm 0.09 $ \\

    Distr. C & $98.54\%$ $\pm 0.1 $ &$98.56\%$ $\pm 0.1 $ &$98.77\%$ $\pm 0.1 $ &$\mathbf{99.0}\%$ $\pm 0.08 $ &$\mathbf{99.0}\%$ $\pm 0.08 $ &$98.45\%$ $\pm 0.08 $ \\

    Distr. A* & $96.56\%$ $\pm 0.27 $ &$13.37\%$ $\pm 3.21 $ &$19.93\%$ $\pm 5.95 $ &$\mathbf{98.09}\%$ $\pm 0.26 $ &$97.99\%$ $\pm 0.26 $ &$96.24\%$ $\pm 0.22 $ \\

    Distr. B* & $99.47\%$ $\pm 0.07 $ &$45.26\%$ $\pm 4.1 $ &$57.51\%$ $\pm 4.02 $ &$\mathbf{99.67}\%$ $\pm 0.09 $ &$99.63\%$ $\pm 0.11 $ &$99.44\%$ $\pm 0.07 $ \\

    CIFAR10\\
    \midrule
    IID (A) & 0.771\% & 0.925\% & $\mathbf{0.927\%}$ & 0.912\% & 0.907\% & 0.776\% \\

    Distr. B & 0.888\% & 0.884\% & 0.85\% & $\mathbf{0.935}$\% & $0.931\%$ & 0.895 \\

    Distr. C & 0.832\% & 0.857\% & 0.868\% & 0.907\% & $\mathbf{0.914}$\% & 0.809\% \\

    Distr. A* & 0.763\% & 0.248\% & 0.259\% & 0.877\% & $\mathbf{0.9}$\% & 0.78\% \\

    Distr. B* & 0.876\% & 0.782\% & 0.476\% & $\mathbf{0.937}$\% & 0.926\% & 0.878\% \\

    \bottomrule
  \end{tabular}
\end{table}

\section{Limitations and Future Work}
The benchmark has some limitations that should be addressed in future work and are discussed in detail in \Cref{appendix limitations}.
Firstly, the theoretical properties of \texttt{WAFFLE} should be investigated, in particular whether the guarantees of \texttt{SCAFFOLD} are maintained.
Secondly, our benchmark should be expanded to cover more types of non-IID distributions and naturally partitioned datasets, and to include more personalized FL methods like the recently published \texttt{Ditto}.
Finally, the versatility of \texttt{WAFFLE} could also be further investigated by optimizing the functions $\Omega$ and $\Psi$ for different contexts.

\section{Conclusion}
In this work, we propose \texttt{WAFFLE}, a personalized FL algorithm based on \texttt{SCAFFOLD}, that can overcome client drift and accelerate convergence to an optimal personalized model. \texttt{WAFFLE} uses the Euclidean distance between agent updates to weigh their contributions and thus transition gradually from global to more local training. We explored the performance of \texttt{WAFFLE} in two cases of non-identical data---concept shift and label skew---on two standard image datasets---MNIST and CIFAR10. We  demonstrate that in most cases \texttt{WAFFLE} shows a faster convergence and a possible improvement in accuracy compared with other personalized FL methods. We also demonstrate that \texttt{WAFFLE} is easier to handle than most personalized learning methods. Indeed, a fixed default value for its single hyperparameter yields competitive results in all test cases. However, by design, \texttt{WAFFLE} can be adapted to a task by modifying $\Omega$ and $\Psi$ functions and thus potentially reach even better accuracy. 

\section*{Acknowledgements}
We thank David Roschewitz for sharing with us his implementation of \texttt{APFL}, and Freya Behrens for inspiring us on how to create synthetic concept shift. We are grateful to Celiane De Luca for her suggestions and comments on the writing of this paper. We thank Vincent Yuan and Damien Gengler for their collaboration with Martin Beaussart on adapting \texttt{Weight Erosion} to a neural network for the first time, which lead directly to the present work \citep{ML4Science_Gengler_Beaussart_Yuan}.

\selectlanguage{english}
\typeout{}
\bibliography{references}
\newpage

\appendix
\section*{Appendix}
\section{Limitations and Future Work} \label{appendix limitations}

\begin{description}
    \item[Non-IID distributions.] This work only focuses on certain categories of non-identical data, which limit our evaluation of \texttt{WAFFLE}. Further categories like Prior probability shift, Concept drift, and Covariate shift should be tested.
    \item[Dataset selection.] The benchmark could also have been more realistic with data sets created for this purpose, like FEMNIST or real world data sets, which would benefit from private personalised learning such as medical images.
    \item[Hyperparameters.] Hyperparameters were also a limitation, as some can be very sensitive (\texttt{Weight Erosion} and \texttt{WAFFLE}) and more rigorous testing methods are necessary to find the optimal settings.
    \item[Computational cost.] A major limitation was the number of runs we were able to perform due to computational time limitations, especially for CIFAR10. This only allowed us to report the results of one seed which reduces the confidence of the results. 
    \item[$\Omega$ and $\Psi$.] The \texttt{WAFFLE} benchmark is also limited by the functions we chose for $\Omega$ and $\Psi$. Both of these functions were designed to offer versatility in \texttt{WAFFLE}, so that they could be optimized for various contexts. Depending on the category of non-identical data, it might be that some type of $\Omega$ and $\Psi$ functions gives better results in general. Therefore, more research should be done to find better functions depending on the problem.
    \item[Benchmark.] \texttt{WAFFLE} must still be benchmarked against more personalized FL methods. This work only shows two methods (\texttt{Weight Erosion} and \texttt{APFL}). However, the performance of \texttt{APFL} was lower than anticipated. It is possible that this is because \texttt{APFL} adapts poorly to the tasks proposed in our setting and further experimentation is recommended.
    \item[Robustness.] Finally, we assume honest participation of agents which may not be the case in reality. While tests should be done to verify the strength of \texttt{WAFFLE} against a data poisoning attack, we assume that \texttt{WAFFLE} should not be vulnerable by design as the weighting mechanism should mitigate vulnerability by ignoring malicious participants.
\end{description}

\section{Figures, Tables and Algorithms}
\label{appendix:figures tables algorithms}
\begin{table}[htp]

  \centering
  \caption{Repartition of labels for the distribution C between our ten agents }
  \label{table:distributionC}
  \begin{tabular}{p{1.3cm} p{0.7cm} p{0.7cm} p{0.7cm} p{0.7cm} p{0.7cm} p{0.7cm} p{0.7cm} p{0.7cm} p{0.7cm} p{0.7cm}}
    \toprule
    \multicolumn{11}{c}{Label}                   \\
    \cmidrule(r){2-11}
     & 0 & 1 & 2 & 3 & 4 & 5 & 6 & 7 & 8 & 9 \\
    \midrule
    Agent 0 & 0 & 0 & 0 & \textcolor{myYellow}{0.1} & \textcolor{myOrange}{0.2} & \textcolor{myRed}{0.4} & \textcolor{myOrange}{0.2} & \textcolor{myYellow}{0.1} & 0 & 0\\ \\
    Agent 1 & 0 & 0 & \textcolor{myYellow}{0.1} & \textcolor{myOrange}{0.2} & \textcolor{myRed}{0.4} & \textcolor{myOrange}{0.2} & \textcolor{myYellow}{0.1} & 0 & 0 & 0\\ \\
    Agent 2 & 0 & \textcolor{myYellow}{0.1} & \textcolor{myOrange}{0.2} & \textcolor{myRed}{0.4} & \textcolor{myOrange}{0.2} & \textcolor{myYellow}{0.1} & 0 & 0 & 0 & 0\\ \\
    Agent 3 & \textcolor{myYellow}{0.1} & \textcolor{myOrange}{0.2} & \textcolor{myRed}{0.4} & \textcolor{myOrange}{0.2} & \textcolor{myYellow}{0.1} & 0 & 0 & 0 & 0 & 0\\ \\
    Agent 4 & \textcolor{myOrange}{0.2} & \textcolor{myRed}{0.4} & \textcolor{myOrange}{0.2} & \textcolor{myYellow}{0.1} & 0 & 0 & 0 & 0 & 0 & \textcolor{myYellow}{0.1}\\ \\
    Agent 5 & \textcolor{myRed}{0.4} & \textcolor{myOrange}{0.2} & \textcolor{myYellow}{0.1} & 0 & 0 & 0 & 0 & 0 & \textcolor{myYellow}{0.1} & \textcolor{myOrange}{0.2}\\ \\
    Agent 6 & \textcolor{myOrange}{0.2} & \textcolor{myYellow}{0.1} & 0 & 0 & 0 & 0 & 0 & \textcolor{myYellow}{0.1} & \textcolor{myOrange}{0.2} & \textcolor{myRed}{0.4}\\ \\
    Agent 7 & \textcolor{myYellow}{0.1} & 0 & 0 & 0 & 0 & 0 &\textcolor{myYellow}{0.1} & \textcolor{myOrange}{0.2} & \textcolor{myRed}{0.4} &  \textcolor{myOrange}{0.2}\\ \\
    Agent 8 & 0 & 0 & 0 & 0 & 0 & \textcolor{myYellow}{0.1} & \textcolor{myOrange}{0.2} & \textcolor{myRed}{0.4} & \textcolor{myOrange}{0.2} & \textcolor{myYellow}{0.1}\\ \\
    Agent 9 & 0 & 0 & 0 & 0 & \textcolor{myYellow}{0.1} & \textcolor{myOrange}{0.2} & \textcolor{myRed}{0.4} & \textcolor{myOrange}{0.2} & \textcolor{myYellow}{0.1} & 0\\ \\

    \bottomrule
  \end{tabular}
\end{table}

In \Cref{algo:waffle}, changes w.r.t. \texttt{SCAFFOLD} are written in red ink \citep{karimireddy2020scaffold}.

\begin{algorithm}[!htbp]
    \SetKwInOut{ServerIn}{server input}
    \SetKwInOut{ClientIn}{agent $i$'s input}
    \SetKwInOut{Output}{output}
    \SetKw{KwComm}{communicate}
    \SetKwFor{AllClients}{on each agent}{in parallel do}{end on agents}
    \SetKwFunction{CalcWeights}{CalcWeights}
    \caption{\texttt{WAFFLE}: \textbf{W}eighted \textbf{A}veraging \textbf{f}or Personalized \textbf{F}ederated \textbf{Le}arning}
    \label{algo:waffle}
    \ServerIn{ Initial global model \x{} and global control variate \c, global step-size $\eta_{g}$, and index \istar{} of the requesting agent }
    \ClientIn{ Initial local control variate \ci, and local  step-size $\eta_{l}$}
    \Output{ Model \x{} optimized for agent \istar}

    \algoHL{
        $\left( \mathbf{\alpha}^{r-2} , \mathbf{\alpha}^{r-1} \right) \leftarrow ( \frac{1}{N} \mathbf{1}_{N \times 1}, \frac{1}{N} \mathbf{1}_{N \times 1})$
    } \;
    \For{ \emph{round} $r \leftarrow 1$ \KwTo $R$}{
        \KwComm $\left(\mathx, \mathc \right)$ to all agents  $ i \in \mathS $ \;
        \AllClients{ $ i \in \textcolor{\algoHLcolor}{ \mathSet } $ }{
            initialize local model $ \mathyi \leftarrow \mathx $ \;
            \For{ $k \leftarrow 1$ \KwTo $K$}{
                compute mini-batch gradient $g_{i}(\mathyi) $ \;
                $ \mathyi \leftarrow \mathyi -\eta_{l}(g_{i}(\mathyi)) - \mathci + \mathc) $ \;
            }
            $ \mathci^{+} \leftarrow \mathci - \mathc + \frac{1}{K \eta_{l}}(\mathx - \mathyi)$ \;
            $\left(\Delta \mathyi,\Delta \mathci \right) \leftarrow (\mathyi - \mathx, \mathci^{+} - \mathci)$ \;
            $ \mathci \leftarrow \mathci^{+} $ \;
            \KwComm  $\left(\Delta \mathyi,\Delta \mathci \right)$ to the server \;
        }
        \algoHL{
            $(\bar{ \mathbf{\alpha}}^{r}, \mathbf{\alpha}^r ) \leftarrow $ \CalcWeights{$N$, $R$, \istar, $r$, $\{ \Delta \mathyi \}$, $\mathbf{\alpha}^{r-1}$, $\mathbf{\alpha}^{r-2}$} \;
            $\mathbf{\alpha}^{r-2} \leftarrow \mathbf{\alpha}^{r-1}$ and $\mathbf{\alpha}^{r-1} \leftarrow \mathbf{\alpha}^r$ \;
        }
        \label{WAFFLE algo line: aggregation of updates and control variates}
        $\left(\Delta \mathx,\Delta \mathc \right) \leftarrow \sum_{i \in \textcolor{\algoHLcolor}{\mathSet}} \textcolor{\algoHLcolor}{\bar{ \alpha}^{r}_i} \left(\Delta \mathyi,\Delta \mathci \right)$ \;
        $\mathx \leftarrow \mathx + \eta_g \Delta \mathx$ and $\mathc \leftarrow \mathc +  \textcolor{\algoHLcolor}{ \Delta \mathc } $ \;
    }

\end{algorithm}

\begin{table}[htp]
  \centering
  \caption{Hyperparameters for personalized FL methods}
  \label{table:parameters}
  \begin{tabular}{p{2.6cm} p{0.7cm} p{0.7cm} p{0.7cm} p{0.7cm} p{0.8cm} p{0.6cm} p{0.6cm} p{0.6cm} p{0.6cm} p{0.6cm}}
    \toprule
    & & & MNIST & & & & & CIFAR10 \\
    \cmidrule(r){2-6} \cmidrule(r){7-11}
     Distr. & A & B & C & A* & B* & A & B & C & A* & B* \\
    \midrule
    \texttt{WE} $(p_d)$ & 0.006 & 0.0067 & 0.006 & 0.0088 & 0.0088 & 1.1 & 1.1 & 0.9 & 0.6 & 0.6 \\
    \texttt{WAFFLE} $(\Delta\Omega)$ & 3.2 & 3.2 & 3.2 & 3.2 & 3.2 & 3.2 & 3.2 & 3.2 & 3.2 & 3.2  \\
    \bottomrule
  \end{tabular}
  
\end{table}

\begin{figure}
    \centering
    \label{distance_penalty}
    \includegraphics[scale=0.3]{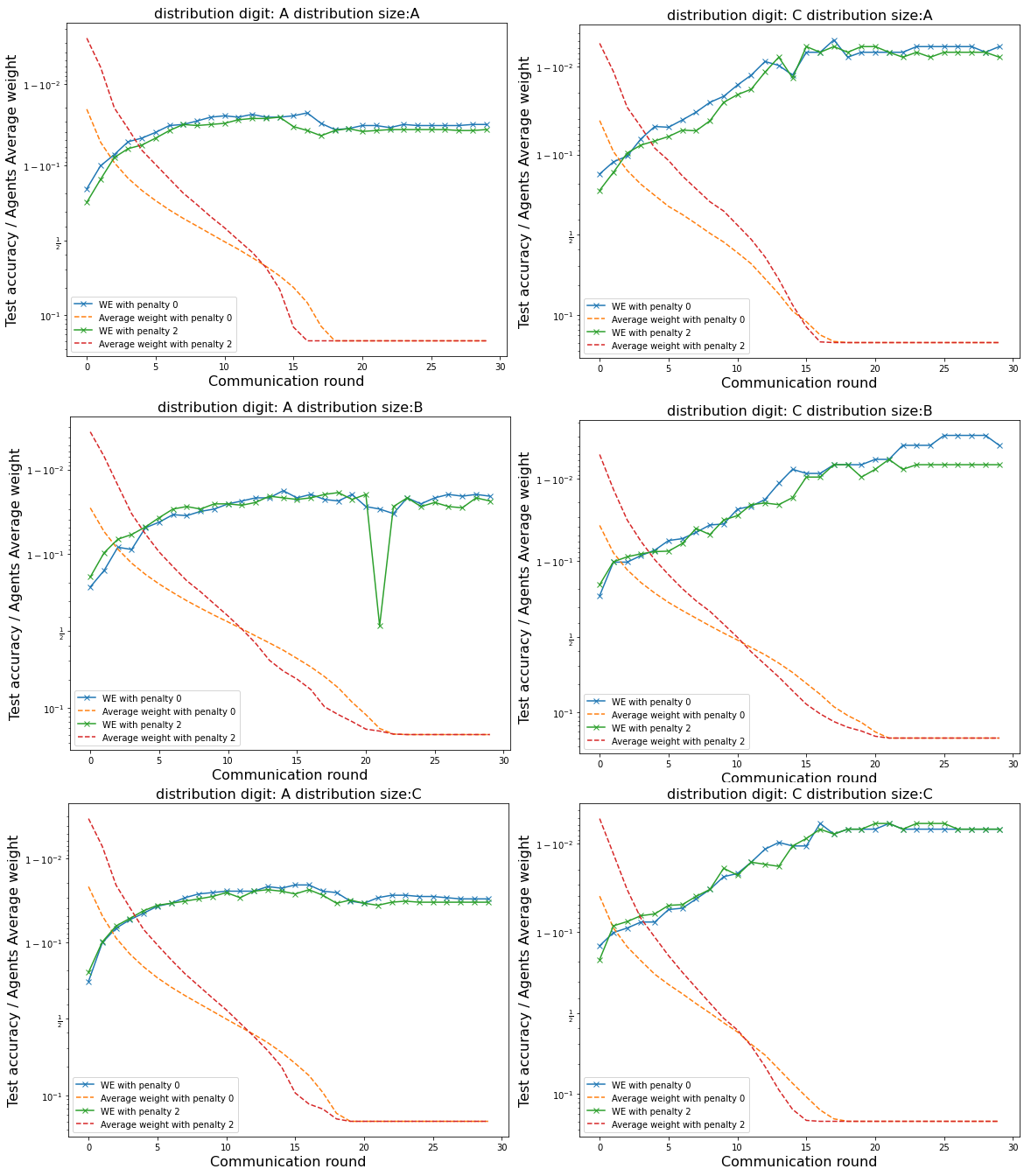}
    \caption{Comparison of the accuracy using MNIST between \texttt{Weight Erosion} with or without a distance penalty factor following different distribution of labels and size of samples. The definition of all distributions used can be seen in \Cref{table:distribution_penalty}. }
\end{figure}

\begin{table}
  \centering
  \caption{Label and size sample distribution for distance penalty experiment, example Label distribution A with Size distribution C mean that agent 4 will have two times more data than agents 0,1,2,3,7,8,9 and data is IID.}
  \label{table:distribution_penalty}
  \begin{tabular}{p{2.5cm} p{0.7cm} p{0.7cm} p{0.7cm} p{0.7cm} p{0.7cm} p{0.7cm} p{0.7cm} p{0.7cm} p{0.7cm} p{0.7cm}}
    \toprule
    \multicolumn{11}{c}{Label}                   \\
    \cmidrule(r){2-11}
     & 0 & 1 & 2 & 3 & 4 & 5 & 6 & 7 & 8 & 9 \\
    Label distribution\\
    \midrule
    A & 0.1 & 0.1 & 0.1 & 0.1 & 0.1 & 0.1 & 0.1 & 0.1 & 0.1 & 0.1\\ \\
    C & 0.25 & 0.25 & 0.25 & 0.25 & 0 & 0 & 0 & 0 & 0 & 0\\ \\
    Size distribution\\
    \midrule
    A & 0.1 & 0.1 & 0.1 & 0.1 & 0.1 & 0.1 & 0.1 & 0.1 & 0.1 & 0.1\\ \\
    B & 0.21 & 0.06 & 0.11 & 0.06 & 0.11 & 0.06 & 0.16 & 0.06 & 0.11 & 0.06\\ \\
    C & 0.1 & 0.1 & 0.1 & 0.1 & 0.2 & 0.05 & 0.05 & 0.1 & 0.1 & 0.1\\ \\
    \bottomrule
  \end{tabular}
\end{table}

\begin{figure}
    \centering
    \includegraphics[scale=0.3]{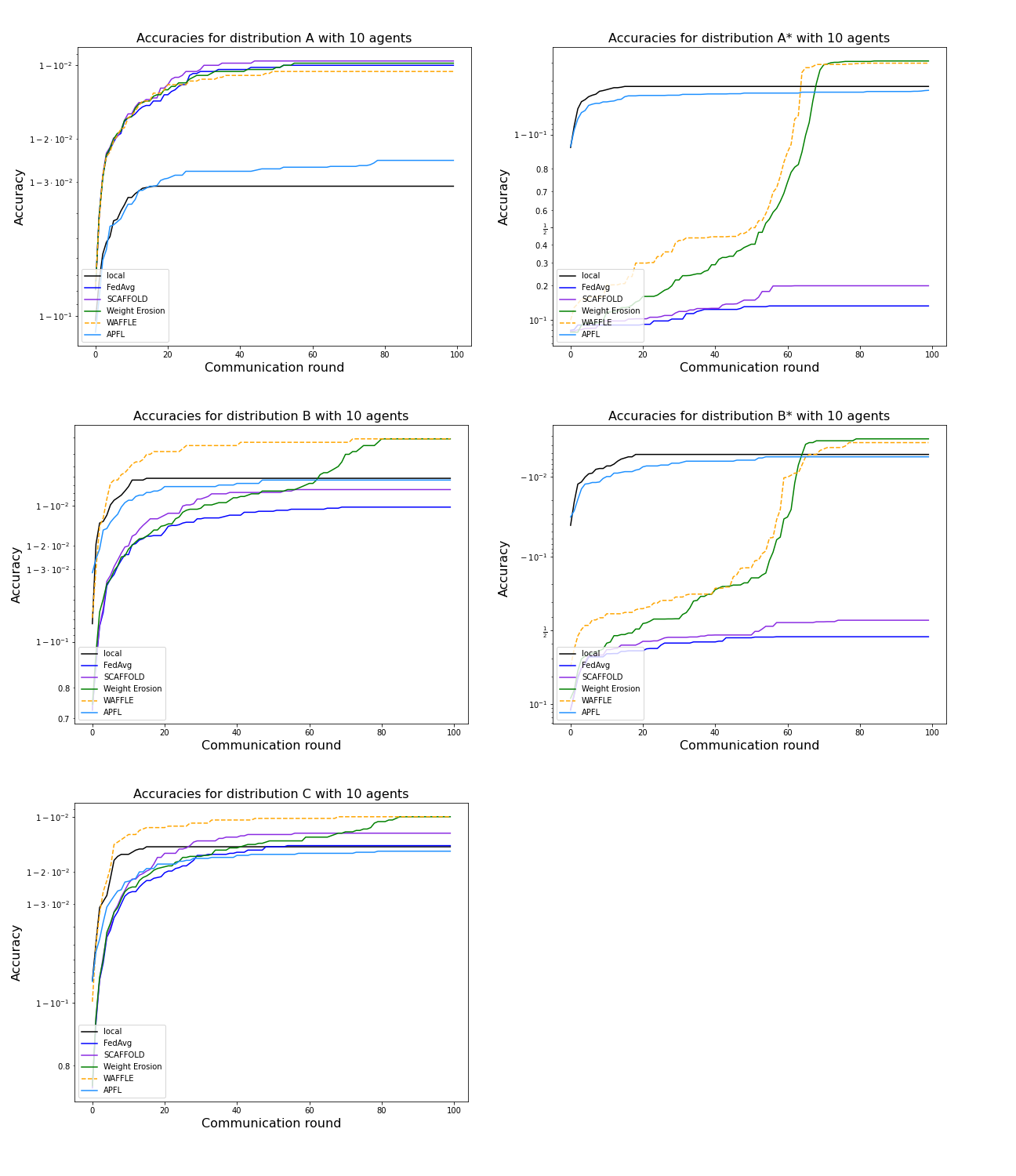}
    \caption{Evolution of the accuracy on MNIST, average over five seeds of the best accuracy obtained up to each turn. }
    \label{acc_mnist}
\end{figure}

\begin{figure}
    \centering
    \includegraphics[scale=0.3]{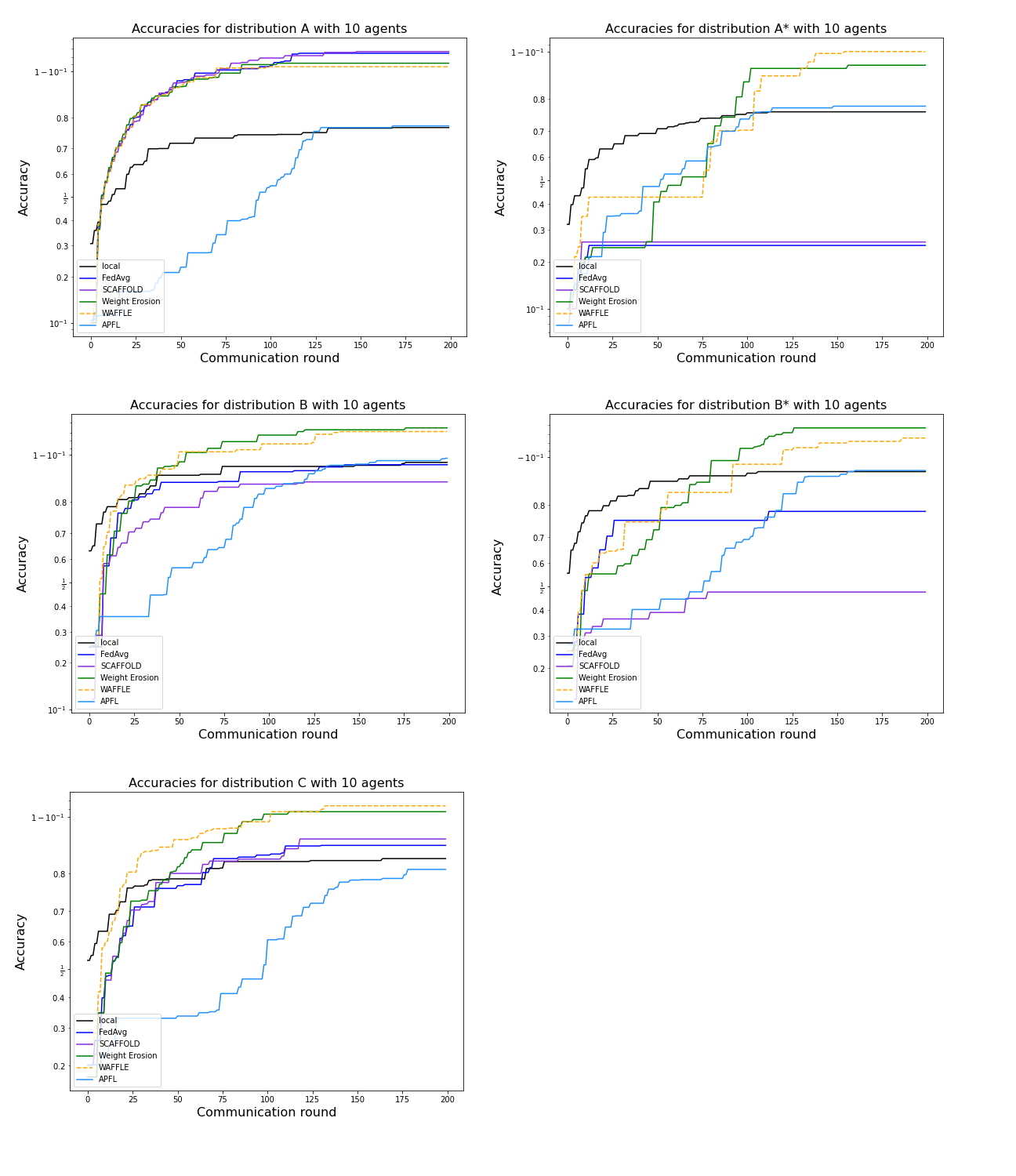}
    \caption{Evolution of the accuracy on CIFAR10, best accuracy obtained up to each turn. }
    \label{acc_cifar}
\end{figure}

\begin{figure}
    \centering
    \includegraphics[scale=0.3]{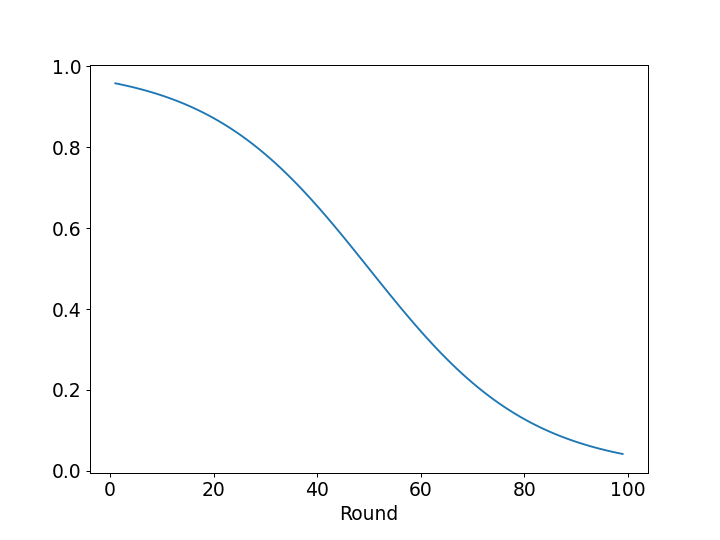}
    \caption{Evolution value of omega  with $\Delta\Omega$  = 3.2}
    \label{omega_curve}
\end{figure}

\end{document}